\pdfoutput=1
\documentclass[runningheads]{llncs}
\usepackage{graphicx}
\usepackage{amsmath,amssymb} 
\usepackage{color}
\usepackage[width=122mm,left=12mm,paperwidth=146mm,height=193mm,top=12mm,paperheight=217mm]{geometry}
\usepackage{times}
\usepackage{epsfig}
\usepackage{graphicx}
\usepackage{amsmath}
\usepackage{amssymb}
\usepackage{epstopdf}
\usepackage{multirow}
\usepackage{subfig}
\usepackage{algorithm}
\usepackage{algorithmic}
\usepackage{cite}
\usepackage{tabularx, booktabs}
\usepackage{arydshln}

\newcolumntype{C}[1]{>{\centering\let\newline\\\arraybackslash\hspace{0pt}}m{#1}}
\newcolumntype{L}[1]{>{\raggedright\let\newline\\\arraybackslash\hspace{0pt}}m{#1}}

\usepackage{fancyheadings}

\usepackage{array,url,kantlipsum}
\newcommand{\Mark}[1]{\textsuperscript{#1}}
\newcommand\blfootnote[1]{%
  \begingroup
  \renewcommand\thefootnote{}\footnote{#1}%
  \addtocounter{footnote}{-1}%
  \endgroup
}

\usepackage[hang,flushmargin]{footmisc}

\newcolumntype{Y}{>{\centering\arraybackslash}X}
\begin{document}
\pagestyle{headings}
\mainmatter
\def\ECCV18SubNumber{1786}  
%
\title{Model-free Consensus Maximization\\ for Non-Rigid Shapes } 
\titlerunning{Model-free Consensus Maximization for Non-Rigid Shapes }
\authorrunning{Thomas Probst, Ajad Chhatkuli, Danda Pani Paudel, and Luc Van Gool}
\author{Thomas Probst\Mark{1}, Ajad Chhatkuli\Mark{1}, Danda Pani Paudel\Mark{1}, and Luc Van Gool\Mark{1,2}}
\institute{	\Mark{1}\,Computer Vision Lab, ETH Z\"urich, Switzerland \\
		\Mark{2}\,VISICS, ESAT/PSI, KU Leuven, Belgium \\
	\email{ \{probstt,ajad.chhatkuli,paudel,vangool\}@vision.ee.ethz.ch}}
\maketitle
%
%
%
%
\begin{abstract}
Many computer vision methods use consensus maximization to relate measurements containing outliers with the correct transformation model. In the context of rigid shapes, this is typically done using Random Sampling and Consensus (RANSAC) by estimating an analytical model that agrees with the largest number of measurements (inliers). However, small parameter models may not be always available. In this paper, we formulate the model-free consensus maximization as an Integer Program in a graph using `rules' on measurements. We then provide a method to solve it optimally using the Branch and Bound (BnB) paradigm. We focus its application on non-rigid shapes, where we apply the method to remove outlier 3D correspondences and achieve performance superior to the state of the art. Our method works with outlier ratio as high as 80\%. We further derive a similar formulation for 3D template to image matching, achieving similar or better performance compared to the state of the art.
\end{abstract}
%
%
\blfootnote{\textbf{Acknowledgements.} Research was funded by the EU's Horizon 2020 programme under grant No.\ 687757-- REPLICATE and grant No.\ 645331-- EurEyeCase. Research was also supported by  the Swiss Commission for Technology and Innovation (CTI, Grant No.  26253.1 PFES-ES, EXASOLVED).}
\section{Introduction}
\label{sec:intro}
Consensus maximization is a powerful tool in computer vision that has enabled practical applications of highly complex algorithms such as Structure-from-Motion (SfM)~\cite{Hartley2004,Longuet1981,Nister2004} to work despite incorrect measurements and noise. Apart from heuristic strategies such as Random Sampling and Consensus (RANSAC)~\cite{Fischler1981}, globally optimal consensus maximizers~\cite{Chin2016, Speciale2017,Bazin2013,Hartley2009,Bazin2014,Li2009,Zheng2011} have been widely studied for rigid shapes, where there exists a simple analytical transformation between two sets of measurements.
In contrast, such tools have not been explored in earnest for the model-free scenario, where simple analytical transformation models cannot explain the measurements. An important field where model-free approaches are needed is in non-rigid shape registration. Consensus maximization in non-rigid shapes have applications in augmented reality, object animations and shape analysis, among others.

While a large number of works have tackled non-rigid registration problem between images or shapes ~\cite{Cho2014,Collins2014b,Kim2011,Lahner2017,Bernard2017}, little attention has been given to identifying outliers in matching correspondences. A few methods solve the problem in the images of non-rigid shapes~\cite{Pilet2008,Pizarro2012} and between a template shape and an image~\cite{Ngo2016} through locally optimal approaches. The difficulty of assigning a suitable minimal parameter model to non-rigid transformations makes it highly challenging to devise a consensus maximizer.

In this paper, we propose a common framework of seeking consensus in a model-free correspondence set. Our key idea is that despite lacking a model which can explain each instance in a matching set individually, one can consider the agreement between two or more instances using certain rules to formulate constraints. In non-rigid shapes, a rule widely applied for reconstruction and registration is the isometric deformation prior. Isometry implies that the geodesic distances are preserved despite deformations.
Using these theoretical understandings, we provide our contributions in three different aspects. First we show how a model-free consensus maximization problem can be posed as a graph problem and solved as an Integer Program if we have inlier/outlier rules on the matching sets. Such an Integer Program can be solved optimally using a BnB approach. Second, we apply this formulation for removing outliers in non-rigid shape correspondences under the isometry prior. We show that our method can handle as much as 80\% outlier correspondences on isometric surfaces. We provide extensive experiments on several isometric and partial shapes, as well as `loosely' isometric partial inter-subject human shapes, where we obtain results that improve over the state-of-the-art methods. To show the generic nature of the introduced consensus maximizer, we also formulate a  3D template-to-image outlier removal problem using the piecewise rigidity and smoothness prior.
We conduct extensive experiments in order to analyze the behavior of the proposed algorithms and to compare with the state-of-the-art methods.

\section{Related Work}
\label{sec:related}
We briefly highlight the related works that are relevant to non-rigid registration problems. The first problem is that of maximizing consensus between matched 3D surface points in non-rigid 3D shapes using the isometry prior. Isometry is a widely used prior in registration~\cite{Kim2011,Aflalo2016,Vestner2017,Lahner2017,Litany2017} as well as 3D reconstruction~\cite{Salzmann2011,Bartoli2015}. Most non-rigid shape registration methods~\cite{Aflalo2016,Vestner2017,Lahner2017,Litany2017} start with a 3D descriptor such as the SHOT descriptor~\cite{Salti2014} or heat kernels~\cite{Ovsjanikov2010} and establish correspondences between shapes through energy minimization. Others compute the registration by blending conformal maps~\cite{Le2016,Kim2011}. Any such matching method results in good and bad matches. In the following sections, we study how the outlier matches from various methods can be removed in practical cases, including complete, partial and inter-subject scenarios.

3D template-to-image matching is yet another important problem in non-rigid shapes that can be used to localize cameras~\cite{Innmann2016} or for template-based 3D  reconstruction~\cite{Wandt2016,Bartoli2015,ChhatkuliPAMI2016,Ngo2016}. Eliminating outlier matches in such cases is addressed in \cite{Ngo2016} by using a local iterative approach. Most other methods which solve image registration~\cite{Cho2014,Pizarro2012} do not use a 3D geometric prior explicitly. We address the problem of consensus maximization in this setting with piece-wise rigidity and smoothness prior. A recent method~\cite{Bernard2017} solves the combinatorial matching problem with similar constraints but does not focus on the problem of identifying outlier matches.
\section{Background and Theory}
\label{sec:background}
\paragraph{Notations.}
We represent sets and graphs as special Latin characters, e.g., $\mathcal{V}$. We use lowercase Latin letters $i, j$, $k$ or $l$ to represent indices or sets of indices. For example, $\mathcal{V}_i$ is an element of the set $\mathcal{V}$. We write known or unknown scalars also in lowercase letters, such as $z$. We use uppercase bold Latin letters to represent matrices (e.g., $\mathsf{M}$) and lowercase bold Latin letters to represent vectors (e.g., $\mathsf{v}$). We use lowercase Greek letter $\epsilon$ to represent thresholds. We use uppercase Greek letters to represent mappings or functions (e.g., $\Phi$). We use $\lVert . \rVert$ to denote the $\ell_2$ -- norm and $\mid . \mid$ to denote the $\ell_1$ -- norm of a vector or the cardinality of a set. Unless stated otherwise, we write primed letters to represent quantities related to the transformed set.

\subsection{Outliers}
Let $\Phi: \Omega \to \Omega'$ be a transformation function between two spatial domains. $\Phi$ is related to the matching sets
$\mathcal{P}= \{\mathcal{P}_i: \mathcal{P}_i \in \Omega,\ i=1,\dots,p\}$ and $\mathcal{P}'= \{\mathcal{P}'_i: \mathcal{P}'_i \in \Omega',\ i=1,\dots,p\}$. In practice, $\Phi$ may be a rigid or non-rigid transformation function or such transformations followed by camera projection. Each member $\mathcal{P}_i$ corresponds to the member $\mathcal{P}'_{i}$ in the second set. This defines a set of matches $\mathcal{C} \subset \mathcal{P} \times \mathcal{P}'$ that may contain outliers. The outlier set $\mathcal{O}$ is defined with a distance function $\Delta$ as:
\begin{equation}
\label{eq:outliergeneric}
\forall i\in \{1\dots p\},\quad
 \Delta\left(\Phi(\mathcal{P}_i),\mathcal{P}'_{i}\right)\geq \epsilon
\implies i\in \mathcal{O}.
\end{equation}
A correspondence pair $(\mathcal{P}_i, \mathcal{P}'_{i})$, also simply denoted as $i$, is an outlier if the distance between the mapping of $\mathcal{P}_i$ and its correspondence $\mathcal{P}'_{i}$, is greater than the threshold $\epsilon$.

\subsection{Consensus Maximization}
Using the definition of outliers in~\eqref{eq:outliergeneric},
the problem of consensus maximization is defined as the minimization of the cardinality of the set $\mathcal{O}$ for the unknown $\Phi$:
\begin{align}
\label{eq:consensusoriginal}
\begin{split}
&\underset{\Phi}{\text{minimize}}\quad |\mathcal{O}| \\
&\text{subject to}\quad \Delta\left(\Phi(\mathcal{P}_i),\mathcal{P}'_{i}\right)\geq \epsilon\implies i\in \mathcal{O}.
\end{split}
\end{align}
Problem \eqref{eq:consensusoriginal} implies that we wish to find the mapping $\Phi$ which results in the least number of disagreements given by the cardinality of $\mathcal{O}$, in the given matching set $\mathcal{C}$. In rigid SfM related problems, $\Phi$ can be often expressed using a linear or non-linear function with a small fixed number of parameters. This means that equation \eqref{eq:outliergeneric} can be evaluated point-wise\footnote{Although in some cases such as that of the Fundamental Matrix, $\Phi$ cannot be determined point-wise, it can be estimated for a minimal set. Thus, a RANSAC problem can be formulated.} and also estimated using a very small size of point correspondence set, known as the minimal set. There is no doubt that such problems can be efficiently solved using RANSAC and other globally optimal methods highlighted in section \ref{sec:intro}. It should be noted that even if $\Phi$ can be parameterized, very recently problem \eqref{eq:consensusoriginal} was shown to be NP-hard with W[1]-complexity~\cite{Chin2017book,Chin2018}, meaning that solving it optimally is very expensive. We call such a problem, when $\Phi$ can be parameterized (with a reasonably small number of parameters), as model-based consensus maximization. In the sections below we focus on the model-free case. Note that most formulations on consensus maximization are written as maximization of the inlier set cardinality rather than the minimization of the outlier set cardinality. However, these definitions are equivalent and we choose the latter for convenience.

\subsection{Generic Rules-based Consensus Maximization}
\label{sec:genericconsensus}
In contrast to model-based problems, for many applications such as those related to non-rigid shapes, $\Phi$ cannot be represented with a small size of parameters and therefore it cannot be estimated using a minimal point set. As a consequence, $\Delta$ cannot be evaluated point-wise. For example, consider the case when $\Phi$ represents the mapping between the two instances of a non-rigid surface. Such a map may be represented by Free-Form Deformation (FFD)~\cite{Brunet2014,Pizarro2012} or specialized latent space models such as SMPL~\cite{Loper2015} for human body, both requiring a large number of points to fit the latent parameters. In such cases problem~\eqref{eq:consensusoriginal} is impractical to solve in its original form.

Therefore, we offer an alternative consensus maximization formulation which is easier to solve for a special class of problems. A problem belongs to this special class if the sets $\mathcal{P}$ and $\mathcal{P}'$ have a common underlying structure which can be measured using subsets of the match set $\mathcal{C}$, without explicitly computing the transformation function $\Phi$. To obtain a tractable formulation, we define a set of binary variables $\{z\}, z_i \in \{0,1\}$ and $i\in\{1,\dots,p\}$ such that $z_i=1 \Longleftrightarrow i\in \mathcal{O}$. Let a binary valued function $\Theta: (\mathcal{S}_a, \mathcal{S}_b)\to \{0,1\} $ measure the agreement between two small subsets $\mathcal{S}_a$, $\mathcal{S}_b\subset\mathcal{C}$. $\Theta$ evaluates to 1 if the subsets $\mathcal{S}_a$ and $\mathcal{S}_b$ agree up to some threshold $\epsilon$ and 0 otherwise. Then the following is an alternative of the original problem~\eqref{eq:consensusoriginal}:
\begin{align}
\label{eq:genericconsensus}
\begin{split}
& \underset{\{z\}}{\text{minimize}}\quad \sum_i z_i \\
& \text{subject to}\, \\
& \Theta(\mathcal{S}_a,\mathcal{S}_b) = 0\implies \exists (\mathcal{P}_i,\mathcal{P}'_{i})\in \mathcal{S}_a\cup \mathcal{S}_b \ \colon \ z_i=1, \\
& \forall\ (\mathcal{S}_a, \mathcal{S}_b): \mathcal{S}_a\neq \mathcal{S}_b,
\end{split}
\end{align}
The function $\Theta$ can be thought of as a rule which uses priors on the sets $\mathcal{P}$ and $\mathcal{P}'$ to measure the agreement on the matched subsets. The subsets $\mathcal{S}_a$ and $\mathcal{S}_b$ sampled from the match set $\mathcal{C}$, are the minimal sets such that $\Theta$ can be evaluated. Problem~\eqref{eq:genericconsensus} simply means, in case two subsets chosen on the basis of a prior do not agree with each other, \emph{at least one member from the union of those subsets must be an outlier}. This is the key idea of our work. Although solving problem~\eqref{eq:genericconsensus} optimally does not guarantee an optimal solution for problem~\eqref{eq:consensusoriginal}, the latter is a close relaxation of the former. Therefore solving problem~\eqref{eq:genericconsensus} amounts to solving the model-free consensus maximization. Problem~\eqref{eq:genericconsensus} is still a combinatorial problem and is NP-hard. In the next section we give more insights into the problem with a graph structure and provide a globally optimal method for solving it with integer programming.

\section{Consensus Maximization with a Graph}
\label{sec:problem}
\begin{figure}[h]
\centering
\includegraphics[width=0.9\textwidth]{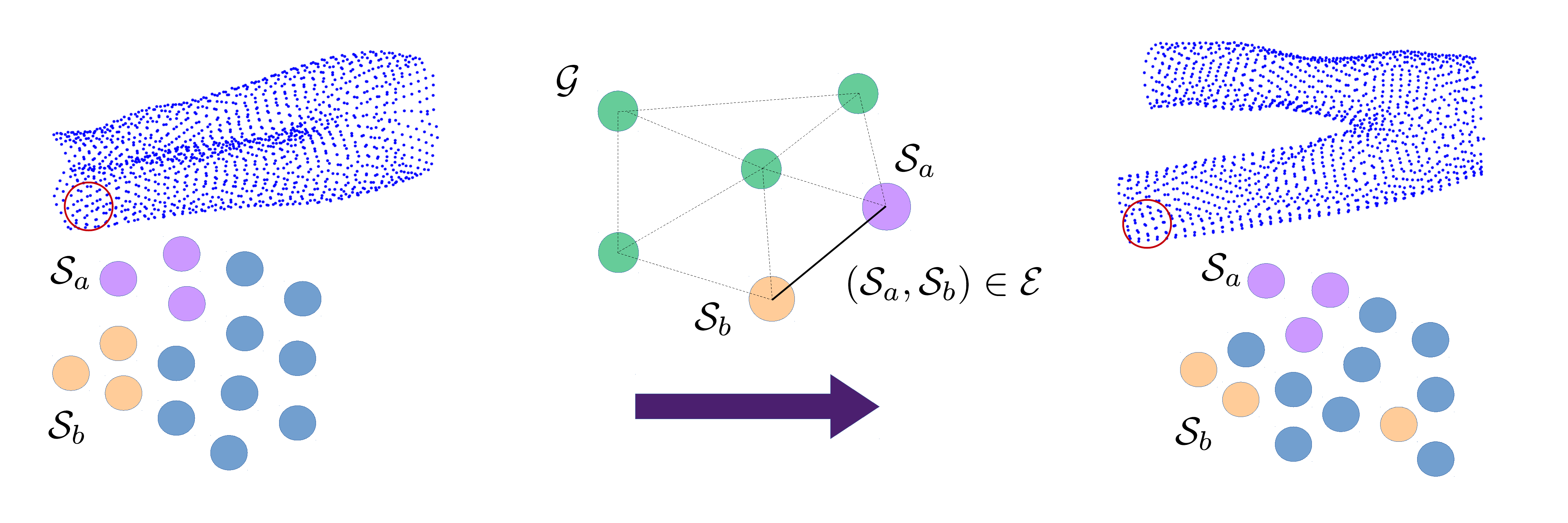}
\caption{\textbf{Graph formulation for consensus maximization.} The selected point sets (nodes) are drawn as orange and purple circles in the graph, connected by edges representing the compatiblity between the sets. The point clouds are taken from \cite{White2007}.}
\label{fig:graph}
\end{figure}

We represent the union of all samples $\mathcal{S}_a$ and $\mathcal{S}_b$ as the nodes and the connection between them as the edges of a graph $\mathcal{G}=\{\mathcal{V},\mathcal{E}\}$. The node set $\mathcal{V}$ consists of all unique sampled subsets $\mathcal{S}_a$ and $\mathcal{S}_b$. An edge $(\mathcal{S}_a, \mathcal{S}_b)\in \mathcal{E}$ connects the nodes $\mathcal{S}_a$ and $\mathcal{S}_b$ and induces the agreement function $\Theta(\mathcal{S}_a, \mathcal{S}_b)$. We use the index $k\in \{1 \dots v\}$ to denote the nodes $\mathcal{V}$ and the index $\ l \in \{1 \dots e\}$ to denote the edges $\mathcal{E}$. Figure~\ref{fig:graph} illustrates this representation of the problem. 

\subsection{Graph Formulation}
Given the graph $\mathcal{G}$, we would still like to compute the original binary variable set $\{z\}$. With a slight abuse of notations, we define the binary variable set of a node as $z_k \triangleq \{z_i\}: (\mathcal{P}_i,\mathcal{P}'_{i})\in\mathcal{V}_k$. Similarly we define the binary variable set of an edge as $z_l \triangleq \{z_i\}:(\mathcal{P}_i,\mathcal{P}'_{i})\in\mathcal{V}_{k_a}\cup \mathcal{V}_{k_b}$ for $\mathcal{E}_l = (\mathcal{V}_{k_a},\mathcal{V}_{k_b})$. The constraint on the binary variables can then be compactly expressed as:
\begin{align}
\label{eq:graphconsensus}
\quad \mathsf{\Sigma} z_l  + \Theta(\mathcal{E}_l) \geq 1,
\end{align}
where $\mathsf{\Sigma} z_l$ represents the sum of all the elements in the set $z_l$. Problem~\eqref{eq:genericconsensus} with constraint~\eqref{eq:graphconsensus} is an example of graph optimization where we need to compute the node properties $z_k$ for each node $k$ using the edge measurements $\Theta(\mathcal{E}_{l})$. 

\subsection{Integer Programming}
Using the constraint of~\eqref{eq:graphconsensus} in a graph, we propose an efficient way to solve the consensus maximization problem, under the framework of Integer Programming, as:
\begin{align}
\label{eq:integerprogram}
\begin{split}
&\underset{\{z\}}{\text{minimize}}\quad \sum z_i \\
&\text{subject to} \quad \sum z_l \geq 1,\quad \forall l\in \{1\dots e\},\quad\text{if}\,\, \Theta(\mathcal{E}_l) = 0.
\end{split}
\end{align}
Problem~\eqref{eq:integerprogram} can be optimally solved using any off-the-shelf solver for Integer Programming. This is done using the popular BnB method. Often such problems in consensus maximization are solved using the so-called big $M$ method~\cite{Mccormick1976}. Such a method is needed when a binary decision function $\Theta$ cannot be defined for a given edge $\mathcal{E}_l$. In that case, the integer inequality in problem~\eqref{eq:integerprogram} is written as $M \sum z_l + \epsilon \geq \Lambda(\mathcal{E}_l)$ using the scalar-valued function $\Lambda$ and a scalar threshold $\epsilon$. Here, $M$ is a chosen large scalar number that makes the problem feasible when $\Lambda$ is large. However, in this work we consider only those problems that can be expressed with a binary rule $\Theta$.

\paragraph{Relaxed alternatives and BnB.}
Integer programming problems are generally non-convex in nature. They can be simplified by further relaxing the binary or integer constraint with real bounds. In contrast, we opt for the BnB approach keeping the integer constraint in order to obtain a globally optimal solution even in case of high outlier ratio. Such an approach computes the lower and upper bound of the cost iteratively and terminates with a certificate of $\epsilon$ sub-optimality if they are equal. We compare the relaxed and the globally optimal methods in section~\ref{sec:results}. In the next section, we describe two different problems in non-rigid shapes which can be expressed in the form of problem~\eqref{eq:integerprogram}.

\section{Non-Rigid Shapes}
\label{sec:app}
Non-rigid objects have deformations that cannot be parameterized with a small fixed set of parameters. Nevertheless, they do obey some shape priors. We provide our methods for two problems in non-rigid shapes below, based on such deformation priors.
\subsection{Shape Matching with Isometry}
\label{sec:shape-matching}
We consider two different shapes $\mathcal{P}$ and $\mathcal{P}'$ related by an unknown deformation $\Phi$. We want to establish the set of outlier points $\mathcal{O}$ on the matching set $\mathcal{C}$. Such problems may arise, for example, when registering 3D non-rigid surfaces using image matches~\cite{Innmann2016} or when registering different shapes with a 3D feature point descriptor~\cite{Salti2014,Ovsjanikov2010}. In order to solve the problem, we consider the isometric deformation prior which assumes that the surface distances are preserved under deformations. The prior allows us to use the following graph attributes:
\begin{align}
\label{eq:shapematching}
\begin{split}
&\mathcal{V}_k = (\mathcal{P}_i, \mathcal{P}'_{i})\\
&\Theta(\mathcal{E}_l)= \begin{cases}
1 \quad & \text{if}\quad \lVert\Psi(\mathcal{P}_{i_1},\mathcal{P}_{i_2}) - \Psi(\mathcal{P}'_{i_1},\mathcal{P}'_{i_2})\rVert \leq \epsilon \\
0 \quad & \text{otherwise}.
\end{cases}
\end{split}
\end{align}
where $\Psi$ is the function that measures the geodesic distance between two points on a surface. Each graph node consists of a single matching pair in $\mathcal{C}$. Therefore each constraint in \eqref{eq:shapematching} obtained for an edge consists of only two binary variables, making the problem  highly sparse. Although, we only show the problem formulation using isometry, other deformation priors such as conformality may be used in problem~\eqref{eq:shapematching}.

\paragraph{Practical considerations.}
While the method works perfectly for isometric surfaces, objects which are undergoing topological changes such as a tearing piece of paper or loosely isometric surfaces such as a human body pose additional difficulty, as isometry is not always satisfied in such cases. We therefore provide a more practical approach to solve the problem in algorithm \ref{alg:shape-registration}. 
\begin{algorithm}[h]
{\small
\caption{\small : $\{z\}$ = shapeRegistration $(\mathcal{P},\, \mathcal{P}',\,\mathcal{C})$}
\label{alg:shape-registration}
\begin{algorithmic}
 \STATE 1. Cluster initial matches $\mathcal{C}$ into $m$ disjoint clusters using k-means.
 \STATE 2. For each cluster $c \in \{1\dots m\}$, \\   
  \hspace{1mm}(a) Compute nearest neighbors and establish edges $\mathcal{E}_l$.\\ 
 \hspace{1mm}(b) For each edge compute the agreement function $\Theta(\mathcal{E}_l)$.\\
 \hspace{1mm}(c) Formulate constraints \eqref{eq:shapematching} with $\Theta$.  
  \STATE 3.  Aggregate all the results from each cluster $c$.
\end{algorithmic}
}
\end{algorithm}
In algorithm \ref{alg:shape-registration}, separately applying the method for different clusters also addresses the non-linear time complexity of the integer programming problem. This allows us to use the method in dense point surfaces as the time complexity with the number of clusters is always linear. To estimate the geodesics, we compute a mesh by Delaunay triangulation when a mesh is not provided.

\subsection{Template to Image Matching}
\label{sec:image-template}
Template-based reconstruction is a well-studied problem~\cite{Bartoli2015,Salzmann2011,ChhatkuliPAMI2016,Ngo2016} which uses matches between the template shape $\mathcal{P}$ and the deformed shape's image $\mathcal{I}$ to reconstruct the deformed surface. Again, the matches established may consist of outliers, in which case the reconstruction obtained can be of poor quality. Here, we propose the use of piece-wise rigidity and surface smoothness as the priors to define the agreement function $\Theta$. Despite non-rigidity, surface smoothness has been successfully used in the state-of-the-art template-based reconstruction methods~\cite{ChhatkuliPAMI2016,Ngo2016}. We use a similar approach by considering that the relative camera to object pose changes smoothly over the surface. Using these priors we define the graph attributes as follows:

\begin{align}
\label{eq:imagematching}
\begin{split}
& \mathcal{V}_{k_1} = \{(\mathcal{P}_{i}, \mathcal{I}_{i})\}, \quad i=\{i_1, i_2, i_3\},\quad i_1, i_2, i_3 \in \{1\dots p \} \\
& \mathcal{V}_{k_2} \in \mathcal{N}(\mathcal{V}_{k_1})\\        
& \Theta(\mathcal{E}_l) = \begin{cases} 1&\quad \text{if}\quad
        \Delta\left(\mathsf{R}^\top_{k_1}, \mathsf{R}_{k_2}\right) \leq \epsilon_1 \, \,
        \text{and} \, \, |\mathsf{t}_{k_1} - \mathsf{t}_{k_2}|
         \leq \epsilon_2 \\
        0&\quad \text{otherwise}\\        
\end{cases}
\end{split}
\end{align}

where $\mathsf{R}_{k_1}$ and $\mathsf{R}_{k_2}$ represent the rotations of the absolute pose estimated using the nodes $\mathcal{V}_{k_1}$ and $\mathcal{V}_{k_2}$ respectively, for the image $\mathcal{I}$. We define $\mathcal{N}(.)$ to be the set valued function giving neighboring nodes in the graph. Similarly $\mathsf{t}_{k_1}$ and $\mathsf{t}_{k_2}$ represent the camera translations. The rule $\Theta$ measures how well the poses agree for the two nodes. To that end, $\Delta$ is the function used to measure the distance between two rotations.  We use two hyperparameters $\epsilon_1$ and $\epsilon_2$ to threshold the change in rotation and translation respectively. Local rigidity and surface smoothness imply that the poses should also change smoothly.
The absolute pose problem can be solved using any of the so-called PnP methods~\cite{Kneip2014,Lepetit2009,Urban2016}. We consider only the minimal problem that uses three non-collinear matched points and is also known as the P3P method~\cite{Kneip2014}. The solutions obtained with P3P have a 4-fold ambiguity. This can be disambiguated either by using an additional matching point pair or by simply choosing the solution that minimizes $\Delta$. The nodes are sampled such that each edge requires only four unique point matches and therefore each inequality constraint will consist of four binary variables.

\paragraph{Practical considerations.}
Piecewise rigidity is a stronger prior compared to isometry. For non-rigid shapes, this holds true only for close neighbors. In contrast to the shape matching problem of \ref{sec:shape-matching}, each edge here requires four point matches. For that reason, it requires the matching set to be dense enough so that the points obey rigidity at least locally. Algorithm~\ref{alg:template-registration} describes the implementation of the method.
\begin{algorithm}[h]
{\small
\caption{\small : $\{z\}$ = templateImageRegistration $(\mathcal{P},\, \mathcal{I},\,\mathcal{C})$}
\label{alg:template-registration}
\begin{algorithmic}
 \STATE 1. Cluster initial matches $\mathcal{C}$ into $m$ disjoint clusters using k-means.
 \STATE 2. For each cluster $c \in \{1\dots m\}$, \\   
  \hspace{1mm}(a) Compute various triangulations of the point clusters and establish edges with two triangles.\\ 
 \hspace{1mm}(b) For each such pair of triangles with shared edge, evaluate $\Theta(\mathcal{E}_l)$.\\
 \hspace{1mm}(c) Formulate constraints \eqref{eq:imagematching} with $\Theta$.  
  \STATE 3.  Aggregate all the results from $m$ clusters.
\end{algorithmic}
}
\end{algorithm}
A very naive simplification of algorithm \ref{alg:template-registration} can be made by considering all points that produce a high number of 1's in the agreement function $\Theta$ to be inliers. We term such a voting method as local filtering which can find obvious inliers in the template-image matching problem.

\subsection{Complexity Analysis}
The combinatorial complexity of problem \eqref{eq:integerprogram} depends on four main aspects: the number of points $p$, the neighborhood size $q$, the cluster size $r$ and the cardinality of the minimal set required to represent a vertex set $\mathcal{S}$, say $s$ in the graph (see Fig.~\ref{fig:graph}). The complexity for a single Integer Program as reported in Table~\ref{tab:complexity} can be directly obtained from the combinatorics in graph.
Although the template-to-image matching complexity ($s=3$) is high, the problem demands only local agreement, which allows us to use a small local neighborhood ($q=15$) for creating the vertices (triangles in this case) with on average $30$ edges per point. This is not the case in the shape matching and we use a fully-connected graph ($q = p/r-1$) on any cluster as the geodesic measurements are valid irrespective of the points' proximity.
\begin{figure}
\centering
\footnotesize
\begin{minipage}[c][2.cm][t]{0.48\textwidth}
\captionof{table}{\textbf{Complexity Analysis.} Solving for $n$ points and minimal set size $s$ with full connectivity (cluster size $r=1$) and $q$-connectivity (cluster size $r$).}
\label{tab:complexity}
\end{minipage}
\hspace{1em}
\begin{minipage}[c][2.cm][t]{0.48\textwidth}
\begin{tabular}{c:c|c:c}      
                  \multicolumn{2}{c|}{\textbf{Full-connectivity}} & \multicolumn{2}{c}{\textbf{$q$-connectivity}}  \\ 
                  Vertices                                      & Edges                                                     & Vertices                                   & Edges                                            \\ \hline
                  $\binom{p}{s}$                                           & $\dbinom{\binom{p}{s}}{2}$                                                                     & $\frac{p}{sr}\binom{q}{s-1}$                                & $\dbinom{\frac{p}{sr}\binom{q}{s-1}}{2}$        
\\ 
\end{tabular}
\end{minipage}
\end{figure}
%
\section{Experimental Results}
\label{sec:results}
We present the results and analysis of our proposed methods in this section on several standard datasets. We refer to the integer program based methods as \emph{exact} or the proposed method. We also compare with the simplified method where the binary constraints in problem~\eqref{eq:integerprogram} have been relaxed to real, which we refer to as the \emph{relaxed} method.
We compare and use several matching or outlier removal methods. We write the spline-warp based image outlier removal method~\cite{Pizarro2012} as \emph{featds}. We write the graph matching method~\cite{Cho2014} as \emph{maxpoolm}. We test the template-image outlier removal method based on mesh Laplacian~\cite{Ngo2016} as \emph{laplacian}. Apart from these image-based methods we also use shape matching methods. We write the recent deformable shape kernel matching method~\cite{Lahner2017} as KM. We write the deep functional map~\cite{Litany2017} as DFM and the blended intrinsic maps~\cite{Kim2011} as BIM.
We implement our methods in MATLAB with YALMIP\cite{yalmip} and MOSEK\cite{Mosek} for integer programming. Below we describe in detail the experiments for each of the discussed non-rigid registration problems.

\paragraph{Clustering and threshold parameters.}
For some experiments, we apply clustering to handle the high number of point matches. For template-to-image matching and the Hand dataset, the number of point matches is low ($n<200$) and therefore the number of clusters is 1. For the human shapes and the newspaper dataset we choose the number of clusters as 5 based on neighborhood (k-means clustering). Note that the result aggregation is straight forward, since the clusters are disjoint. For Fig.~\ref{fig:method_analysis}, to vary the number of points, we randomly sub-sampled the points to a fixed number.
Regarding thresholds, we fix $\epsilon = 20\%$ distance error relative to the template for shape matching (Sec.~\ref{ssec:shape-matching}) unless stated otherwise. In the template-to-image (Sec.~\ref{ssec:image-template}) matching case, we use $\epsilon_1 = 10^{\text{o}}$ and $\epsilon_2 = 40\%$ for all datasets.

\subsection{Non-rigid Shape Matching}
\label{ssec:shape-matching}
We begin by analyzing the behavior of the proposed methods on synthetic data where the ground truth correspondences are available for the shape matching problem. We also compare the proposed methods with the state-of-the-art methods on several real datasets. All these are outlined below.

\paragraph{Mocap data.}
We test with two cloth-capture data~\cite{White2007}. The datasets consist of a cloth falling (toss) and a moving pair of trousers (stepping trousers). The datasets are generated with mocap and consist of registered real 3D points. We synthetically generate outliers by randomly re-assigning matches to evaluate our methods. 

\begin{figure*}[ht]
\centering
\scriptsize
  \begin{minipage}[t]{0.28\textwidth}
  \centering
   \includegraphics[width=1\textwidth]{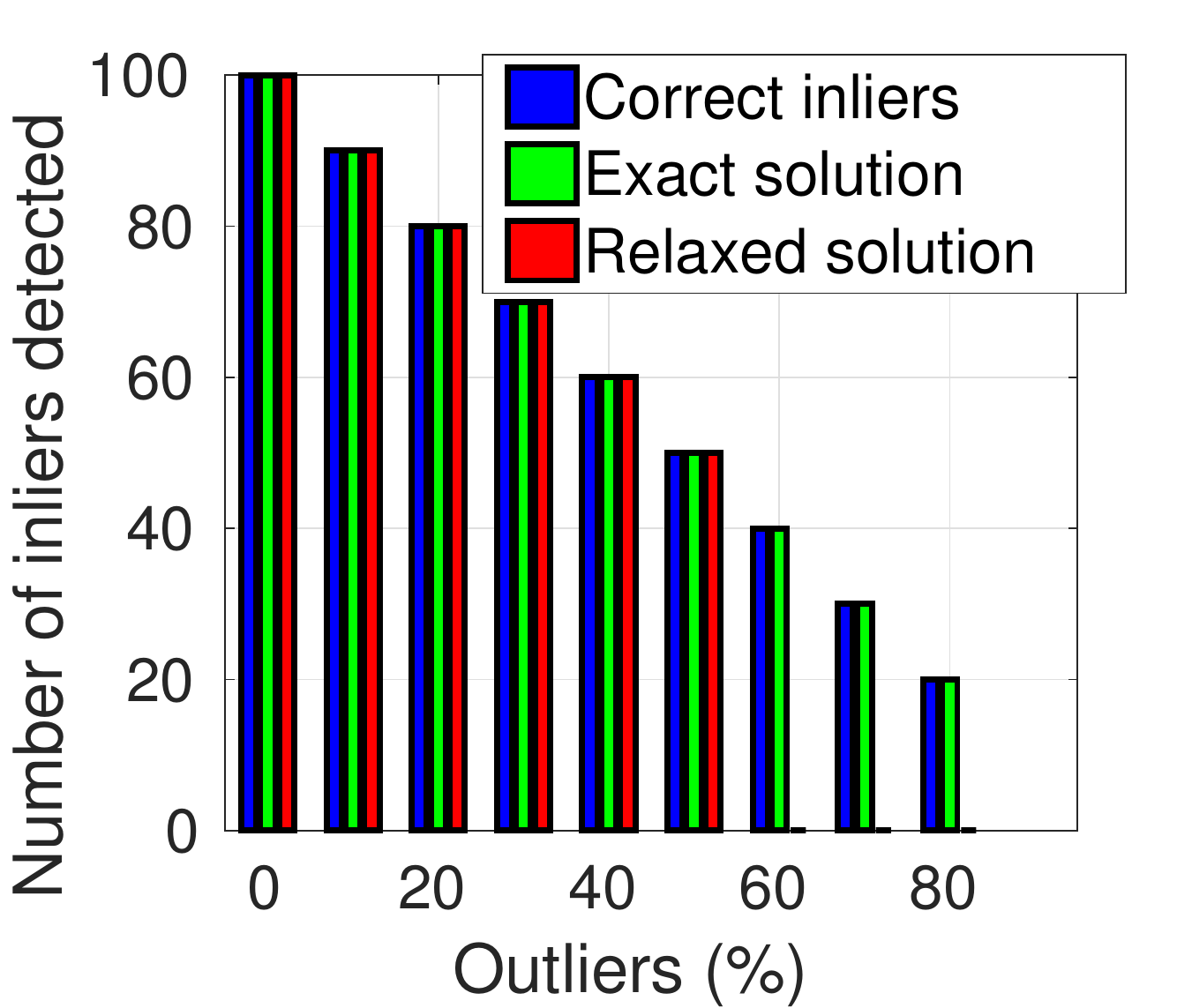}
   (a) \emph{exact} vs. \emph{relaxed}
  \end{minipage}  
  \begin{minipage}[t]{0.28\textwidth}
  \centering
   \includegraphics[width=1\textwidth]{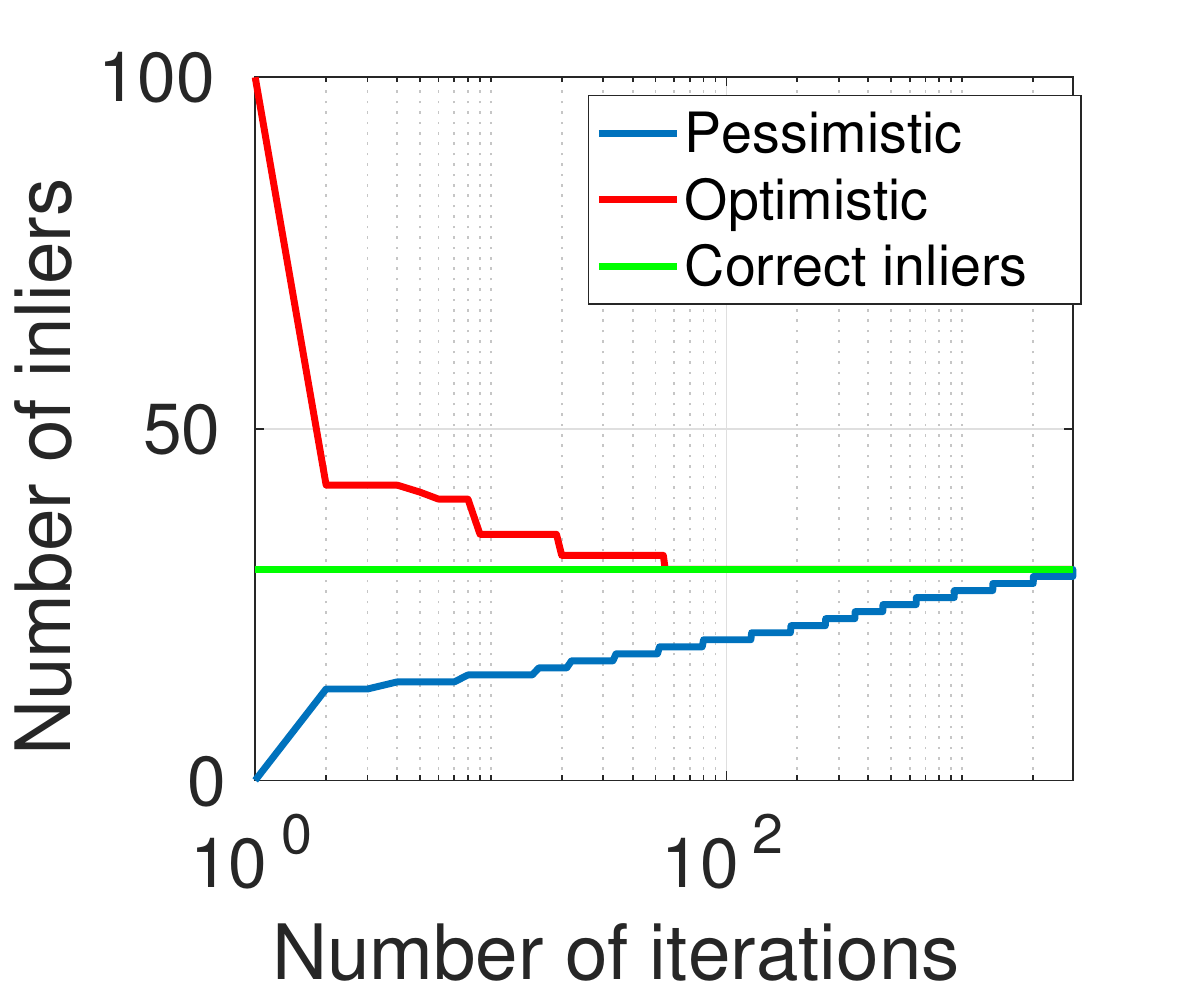}
   (b) BnB Convergence
  \end{minipage}  
  \begin{minipage}[t]{0.28\textwidth}
  \centering
   \includegraphics[width=1\textwidth]{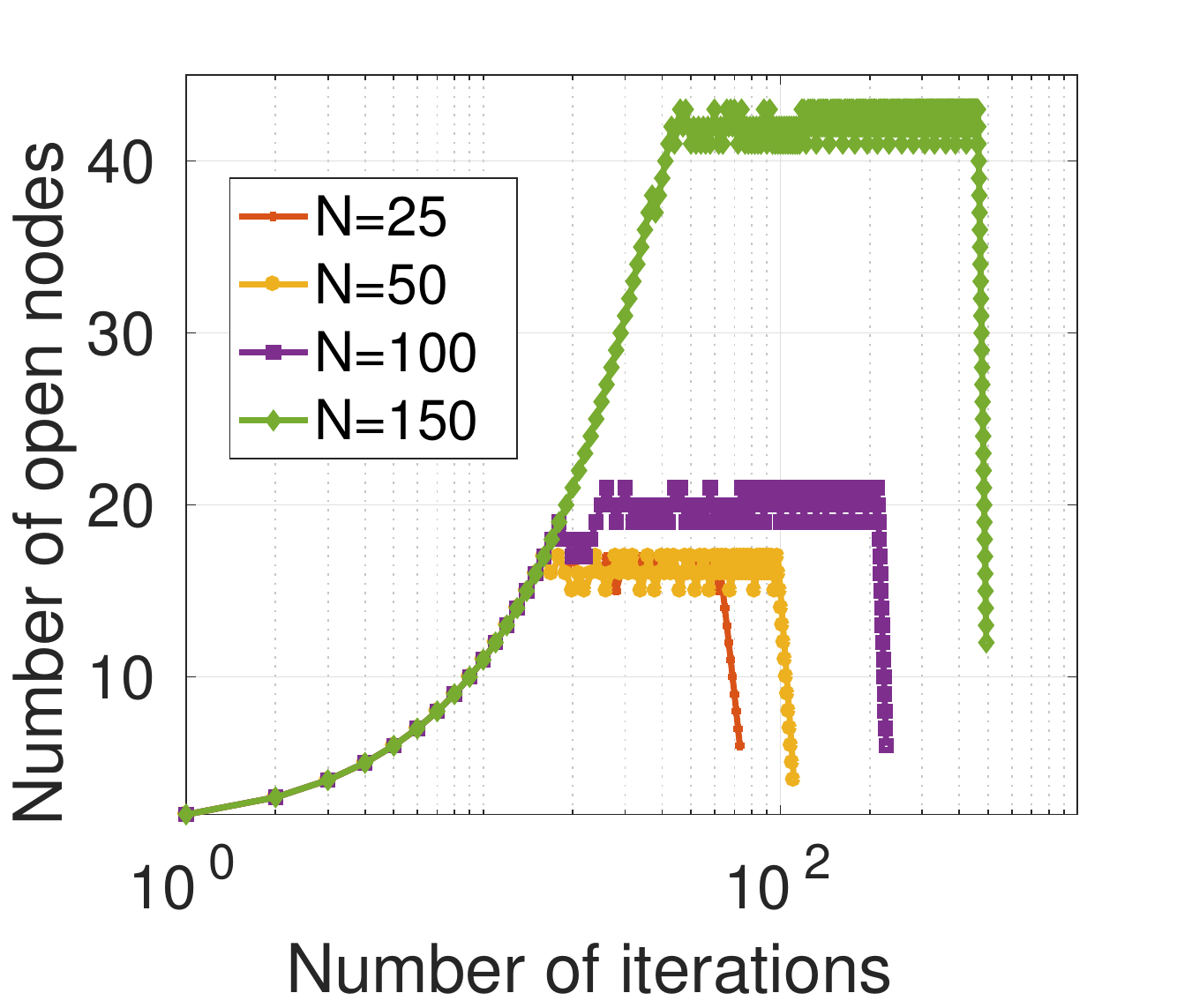}
   (c) BnB open nodes (50\% outliers)
  \end{minipage}
   \includegraphics[width=0.86\textwidth]{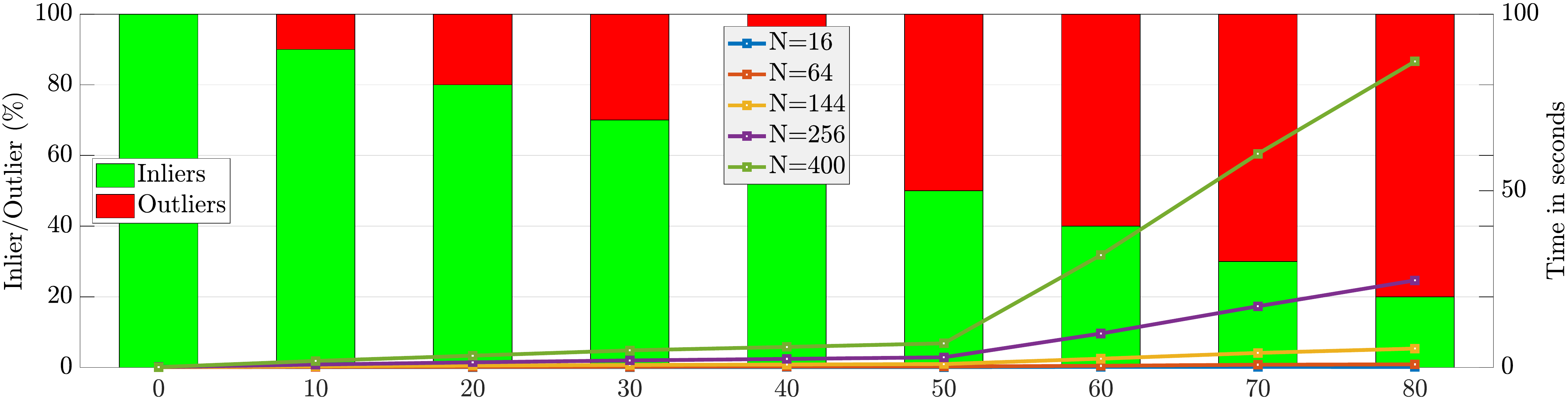}  \\
   (d) Run time of our method with increasing number of points and outlier percentage.
   \caption{\textbf{Analysis of our method.} Number of inliers detected, convergence  of the proposed method, and time taken for the mocap cloth dataset~\cite{White2007} under various setups. Note that the number of iterations in (b) and (c) are in log-scale.}
\label{fig:method_analysis}
\end{figure*}

Figure~\ref{fig:method_analysis} (a) compares the \emph{relaxed} and \emph{exact} versions of the proposed method. We observe that, for low outlier ratio, it is possible to remove all the outliers using the \emph{relaxed} method. However, it breaks down as the percentage of outliers increase beyond 50\%, while the \emph{exact} solution still correctly detects the inliers even in conditions with 80\% outliers. Note that the proposed method does not detect any false positive inliers. Figure~\ref{fig:method_analysis} (b) shows how the \emph{exact} method behaves with the number of iterations. We observe that the method quickly computes the upper bound cost or the pessimistic inlier set while it takes a while to obtain the certificate of optimality. We find this behavior to be consistent to many other experimental setups. 
Figure~\ref{fig:method_analysis} (c) shows the number of open nodes at each iteration, describing how BnB evaluates and prunes branches. 
To investigate time complexity, we also plot the execution time for the \emph{exact} method in figure~\ref{fig:method_analysis} (d). It can be observed that the execution time increases with increase in the number of points. However, this is not a problem in practice thanks to the clustering framework presented in algorithm \ref{alg:shape-registration}.

\begin{figure}
\centering
\small
  \begin{minipage}[t]{0.49\textwidth}
  \centering
   \includegraphics[width=1\textwidth]{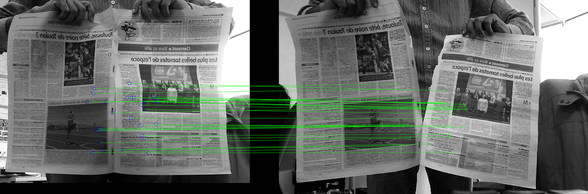}     
  \end{minipage}
  \begin{minipage}[t]{0.49\textwidth}
  \centering   
   \includegraphics[width=1\textwidth]{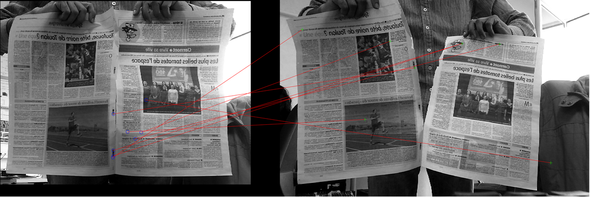}   
  \end{minipage} \\
  (a) \emph{exact}
  
  \begin{minipage}[t]{0.49\textwidth}
  \centering
   \includegraphics[width=1\textwidth]{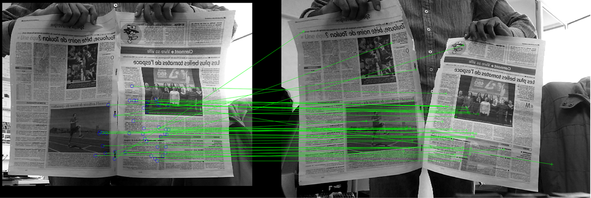}      
  \end{minipage}
  \begin{minipage}[t]{0.49\textwidth}
  \centering
   \includegraphics[width=1\textwidth]{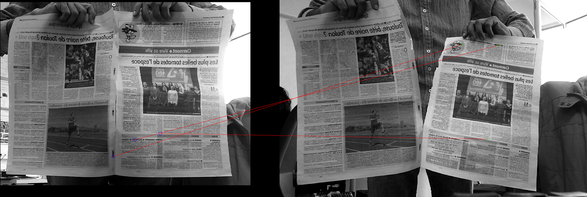}   
  \end{minipage}
  (b) \emph{relaxed}
  
  \begin{minipage}[t]{0.49\textwidth}
  \centering
   \includegraphics[width=1\textwidth]{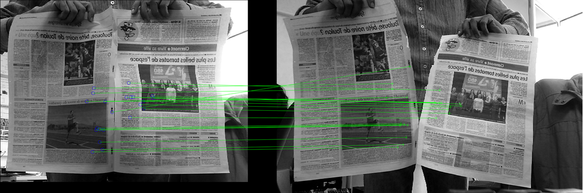}   
  \end{minipage}
  \begin{minipage}[t]{0.49\textwidth}
  \centering
   \includegraphics[width=1\textwidth]{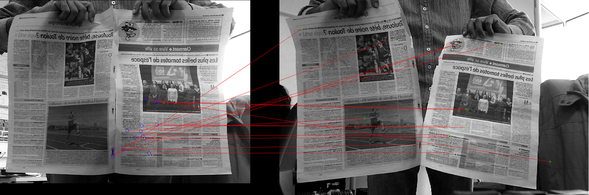}   
  \end{minipage}
  (c) \emph{laplacian}
%
\caption{\textbf{Newspaper dataset.} Visualization of inlier and outlier matches from our \emph{exact} and two next best performing methods for an example pair of the Newspaper dataset. Left column shows the inlier detection and the right column shows the outlier detection.}
\label{fig:newspaper_qualitative1}
\end{figure}

\paragraph{KINECT Newspaper dataset.}
The RGB-D data obtained from depth-camera sensors such as KINECT make an important field of application for the method. We investigated our method on the Newspaper dataset\footnote{\label{fn:dataset} Dataset was provided by the authors.}~\cite{Chhatkuli2016}. It consists of a double sheet of newspaper being torn into two parts. Figure~\ref{fig:newspaper_qualitative1} shows the inliers and outliers for a part of the template image with our method. Due to the local neighborhood computed using both point sets, the \emph{exact} method can robustly handle the topological changes. On the other hand, the \emph{relaxed} method does not work well from lack of enough constraints\footnote{The complete set of results are provided in the supplementary document.}.

\begin{figure}
\centering
\small
  \begin{minipage}[t]{0.49\textwidth}
  \centering
   \includegraphics[height=1.95cm]{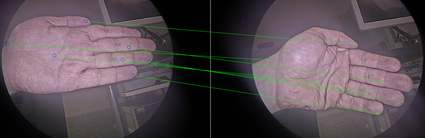}\\
  \end{minipage}
  \begin{minipage}[t]{0.49\textwidth}
  \centering
   \includegraphics[height=1.95cm]{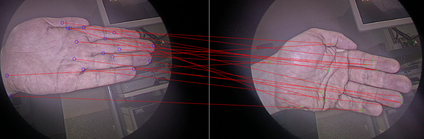}\\
  \end{minipage}
\caption{\textbf{SfM Hand dataset.} Inlier detections (left) and outlier detections (right) of our \emph{exact} method.}
\label{fig:hand_qualitative}
\end{figure}

\begin{figure}
\centering
\small
  \begin{minipage}[t]{0.49\textwidth}
  \centering
   \includegraphics[height=2cm]{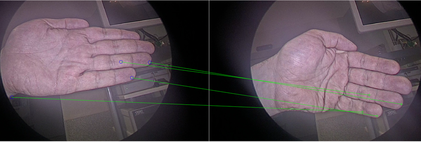}
  \end{minipage}
  \begin{minipage}[t]{0.49\textwidth}
  \centering
   \includegraphics[height=2cm]{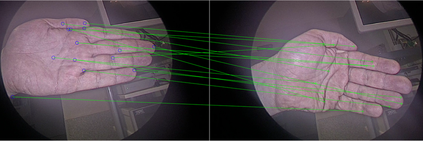}
  \end{minipage}
\caption{Inlier detections with \emph{laplacian} (left) and \emph{relaxed} (right).}
\label{fig:hand_qualitative_others}
\end{figure}

\paragraph{Hand dataset.}
The hand dataset~\cite{Chhatkuli2016} consists of two different instances of a hand and their 3D ground truth obtained with SfM. Due to the non-rigid deformation, the detected SIFT correspondences consist of very few matches with a large percentage (more than 70\%) of outliers. 
The shape matching methods~\cite{Lahner2017,Kim2011} completely fail on this dataset and we do not show them here. We show the results of the \emph{exact} method in figure~\ref{fig:hand_qualitative} and the next best performing methods in figure~\ref{fig:hand_qualitative_others}. These qualitative results clearly show that the compared methods do not perform well in such difficult cases.

\paragraph{Human body shapes.}
In the next set of experiments, we use our methods on human body scans from the FAUST~\cite{Bogo2014} dataset. To introduce challenging outliers, we consider a partial matching scenario by cutting out one arm and one leg from the mesh, and matching it to the full one. Thanks to the mesh registrations provided by the dataset, we can exactly evaluate inliers and outliers based on geodesic deviations to the ground truth correspondences (deviations greater than 15cm are considered as outliers). We compare our \emph{relaxed} and \emph{exact} methods against matches estimated by DFM~\cite{Litany2017}, KM~\cite{Lahner2017}, and BIM~\cite{Kim2011}. Although BIM~\cite{Kim2011} produced visually good correspondences, it suffered from mirror-image ambiguity, that could not be resolved. Therefore we compare to BIM only where proper evaluations were possible.  

\begin{figure*}
\scriptsize
\centering
\begin{tabular}{L{2.25cm}||C{2cm}:C{1.cm}|C{2cm}:C{1.cm}|C{2cm}:C{1.cm}}
\multicolumn{1}{l}{\multirow{ 4}{*}{\textbf{Method}}} \\ 
&\multicolumn{4}{c|}{FAUST} &\multicolumn{2}{c}{MPII HumanShape}\\
&\multicolumn{2}{c}{\emph{intra-subject}} & \multicolumn{2}{c|}{\emph{inter-subject}}&\multicolumn{2}{c}{\emph{from rendered depth map}}\\
			& Inliers / Outliers & Time [s] 	& Inliers / Outliers  & Time [s] 		& Inliers / Outliers  & Time [s]  \\
\hline
BIM 			& - & -				& 3381 / 1602 & 3		& - & - \\
BIM+Ours (\emph{relaxed})   	& - & - 			& 3269 / \textbf{1362} & 10	& - & - \\
BIM+Ours (\emph{exact})   	& - & - 			& 3267 / 1395 & 32.9		& - & - \\
\hline
DFM 			& 4211 / 772 & 1			& 3756 / 1227 & 1			& 272 / 3728	& 1\\
DFM+Ours (\emph{relaxed})   	& 3918 / \textbf{31} & 19		& 3437 / \textbf{93} & 15		& -		&-\\
DFM+Ours (\emph{exact})   	& 3918 / \textbf{31} & 24		& 3437 / \textbf{93} & 19.4		& -		&- \\
\hline
KM			& 4736 / 181 & 89 		& 4051 / 860 	   & 92		& 572 / 3387& 53\\
KM+Ours (\emph{relaxed})   	& 4554 / 18 & 104		& 3634 / \textbf{161}& 107	& -& - \\
KM+Ours (\emph{exact})   	& 4556 / \textbf{17} & 110 	& 3634 / \textbf{161}& 115	& -& - \\
\hline
RF			& - & -				& - & -		& 3220 / 780& $<$1\\
RF+KM		  	& - & - 			& - & -		& 1162 / 269& 3\\
RF+Ours (\emph{relaxed})   	& - & - 			& - & -		& 2800 / \textbf{137}&14\\
RF+Ours (\emph{exact})   	& - & -				& - & -		& 2800 / \textbf{137}&15\\
RF+ICP  	  	& - & - 			& - & -		& 3166 / 159 &301\\
\hline
& \multicolumn{2}{c|}{mostly isometric} &\multicolumn{4}{c}{non-isometric}\\
\end{tabular}
\captionof{table}{\textbf{Non-rigid 3D shape matching.} Results on FAUST~\cite{Bogo2014} intra- and inter-subject, as well as matching depth maps to the MPII~HumanShape~\cite{Pishchulin2017} model. We report the number of true positive (inliers) and false positive (remaining outliers) matches. \label{tab:shapeMatch}}

\includegraphics[width=1\textwidth]{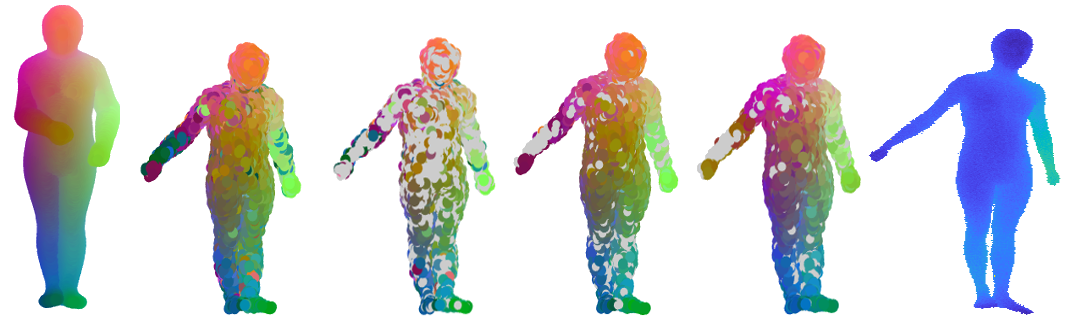}
\captionof{figure}{\textbf{Qualitative results.} Non-isometric shape matching from depth map. Left to right: body mesh model~\cite{Pishchulin2017}, RF~\cite{Pons-Moll2015}, RF+KM~\cite{Lahner2017}, RF+Ours, RF+ICP, input depth map. Correspondences are color-coded, gray indicates removed matches. \label{fig:bodies}}
\end{figure*}

Since our method is designed for isometric shapes, we conduct the first experiment in the \emph{intra-subject} case (same subject in 9 different poses). We observe that our method can successfully eliminate more than 90\% outliers produced by DFM and KM while removing only a few true inliers, as shown in the first column of Table~\ref{tab:shapeMatch}.

In inter-subject body shape matching applications however, the isometry assumption holds only to some extent. We use two challenging datasets to test such scenarios. The first one is on \emph{inter-subject} matching on the FAUST data, again in the partial matching setting. Since the body shape varies across subjects, isometry doesn't hold anymore. The results presented in the second column of Table~\ref{tab:shapeMatch} demonstrate that this problem is significantly harder than the isometric matching. We see that the matches resulting from BIM contain outliers that are very hard to detect, and only 15\% can be removed without sacrificing many inliers. For DFM and KM, we reliably detect more than 80\% and 90\% resp., and therefore improve the matching robustness for subsequent tasks.

Our third experiment with human body shapes involves dense correspondence estimation from a depth map to the 3D model. We rendered synthetic depth map mimicking the projection and noise properties of KINECT from an articulated MPII Human Shape model~\cite{Pishchulin2017} using variations of upright poses and body shapes. To compute the geodesics on this modality, we triangulated the point cloud using 2D Delaunay triangulation. Applying DFM and KM on the raw input does not work well, since SHOT~\cite{Tombari2010} and HKS~\cite{Sun2009} are not reliable features for depth maps. We therefore take initial matches from a metric regression forest~\cite{Pons-Moll2015} trained on the specific task of dense correspondence estimation. We then compare our methods, KM and ICP on top of these matches in the third column of Table~\ref{tab:shapeMatch}. We can conclude that, although provided with inital matches, KM fails to correctly match the two modalities. Our method however shows promising results even though the shapes are non-isometric, and geodesics are computed on the triangulated point cloud. Interestingly, our result is comparable to that of the articulated non-rigid ICP which exploits additional information such as the kinematics and a stronger shape prior. Fig.~\ref{fig:bodies} shows a qualitative example from our test set.

In summary, we showed that our method can be used on top of generic matching methods to robustly detect outliers for isometric deformations, and some classes of non-isometric registration such as inter-subject body shapes. Moreover, we can confirm our results on the synthetic data and conclude that even the \emph{relaxed} method provides good results if the proportion of the outliers is below 50\%.

\begin{table}
\small
\centering
\begin{tabular}{|C{2cm}|C{1.5cm}:C{1.5cm}|C{1.5cm}:C{1.5cm}|C{1.5cm}:C{1.5cm}|}
\hline
\multirow{ 1}{*}{Method}  &\multicolumn{2}{c|}{Kinect Paper} & \multicolumn{2}{c|}{T-shirt}&\multicolumn{2}{c|}{Sintel}\\
  & Inliers & Time(s) &Inliers & Time(s) &Inliers & Time(s)  \\
\hline
Local-filtering  & 46 / 142 & 4.22& 95 / 351 &6.10&17 / 68&2.03\\
\hline
\emph{relaxed}   & 99 / 142 & 5.56 & 291 / 351 & 7.52& 44 / 68& 3.51\\
\hline
\emph{exact}      & 114 / 142 & 7.59 & \textbf{309} / 351 & 9.66& \textbf{53} / 68&5.01\\
\hline
\emph{laplacian}		& \textbf{126} / 142 &	\textbf{1.15} & 301 / 351 & 7.84 &  44 / 68  & 0.53 \\
\hline
\emph{featds}		& 76 / 142 &	3.93	  & 304 / 351 & \textbf{1.46} &  42 / 68  & \textbf{0.32} \\
\hline
\emph{maxpoolm}		& 3 / 142 &	159.96	  & 6 / 351 & 608.55 & 16 / 68  & 7.88 \\
\hline
\end{tabular}
\caption{\textbf{3D template to image matching.} Comparison on three different real datasets.\label{tab:withOutliersComparision}}
\end{table}

\subsection{Template to Image Matching}
\label{ssec:image-template}
The template 3D to image matching is an important problem in non-rigid geometry. Most reconstruction methods~\cite{Ngo2016,ChhatkuliPAMI2016} are sensitive to outlying correspondences and proceed by first removing outliers in matches.
We use problem~\eqref{eq:imagematching} to formulate the template to image outlier removal method with the help of piece-wise absolute pose. We test our results on three datasets: KINECT Paper~\cite{Varol2012}, T-Shirt~\cite{ChhatkuliBMVC2014} and the MPI Sintel~\cite{Butler2012} all of which contain the groundtruth 3D data and images.
The KINECT Paper consists of VGA resolution RGB-D images of a large piece of paper smoothly deforming over time. The T-Shirt data consists of high-resolution wide-baseline images and 3D of a deforming t-shirt. The Sintel data is an animated movie with groundtruth depth. We select a random single pair for each dataset and compute the SIFT matches. We count the number of inliers and outlier matches manually for each of the methods' output.
We compare our methods with three other state-of-the-art methods \emph{laplacian}, \emph{featds} and \emph{maxpoolm}. Similarly, as discussed in section \ref{sec:image-template} we also report the results of the \emph{relaxed} method. We further report the results of the local-filtering method as another baseline where the inliers are decided based on the local neighborhood voting.    

\begin{figure}
\centering
    \begin{minipage}[t]{0.49\textwidth}
  \centering
   \includegraphics[height=1.8cm]{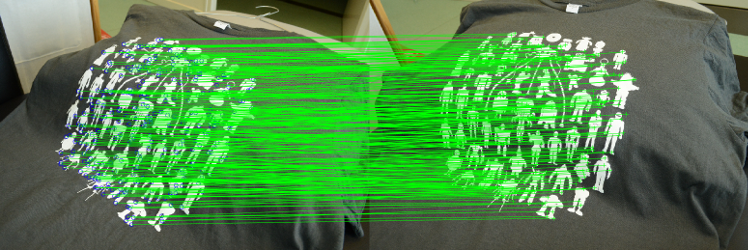}
  \end{minipage}
  \begin{minipage}[t]{0.49\textwidth}
  \centering
   \includegraphics[height=1.8cm]{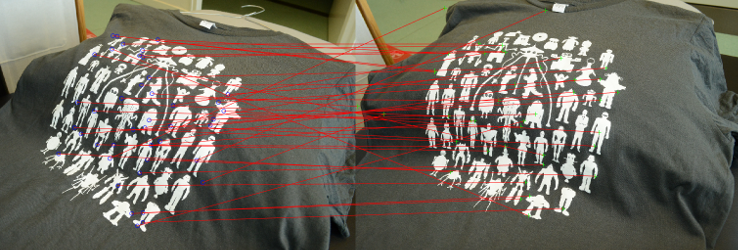}
  \end{minipage}
   \caption{Inliers (left) vs.\ Outliers (right) for the T-shirt dataset using the \emph{exact} method. The performance of our method is on average better than that of the two compared methods designed for non-rigid matching. More results are provided in the supplementary material.}
   \label{fig:tshirt-abspose}
\end{figure}

We test all the methods with favorable parameters. The reported inliers are manually validated. The results show that our method performs in par with \emph{laplacian} designed exactly for the template-based outlier removal. Note that the \emph{exact} method consistently detects more number of inliers than other methods. Our method performs better than \emph{featds} in multi-body situation as \emph{featds} uses a single spline-based warp and computes the residuals to identify outliers. We visualize the results of outlier removal in figure \ref{fig:tshirt-abspose} for the proposed method and two other best performing methods: \emph{featds} and \emph{laplacian}.

\vspace{-0.25cm}
\section{Conclusions and Future Work}
\label{sec:conclusion}
In this paper we brought forward a theory on model-free consensus maximization using integer programming and an optimal method to solve it using Branch and Bound. We formulated two different registration problems using our consensus maximizer: isometric shape outlier removal and template-image outlier removal. We obtained very good results at up to 80\% mismatches in non-rigid shape registration and $>$25\% mismatches in template-image registration. We obtained these results by solving a close relaxation of the original problem with guaranteed optimality. We showed with extensive experiments that our methods consistently performs on par or better than the existing methods. 

Although the focus of this paper was on non-rigid shapes, many vision problems can be converted to formulation~\ref{eq:integerprogram} with three or less variables per graph node. A non-exhaustive list includes: \emph{i)} one variable problems: relative pose on robot navigation~\cite{Scaramuzza2011}, \emph{ii)} two variable problems: robust triangulation~\cite{Li2007} and pure translation estimation~\cite{Fredriksson2014}, and \emph{iii)} three variable problems: image to image affine homography and three-view modulus constraints~\cite{Polleyfeys1999}. For future works, we intend to tackle some of these problems using the model-free consensus maximizer we developed in this paper.

%
{\small
\bibliographystyle{splncs}
\bibliography{egbib}
}
\end{document}